\documentclass[10pt,twocolumn,letterpaper]{article}

\usepackage[pagenumbers]{cvpr}      

\usepackage{colortbl}
\usepackage{graphicx}  
\usepackage{amsmath}
\usepackage{amssymb}
\usepackage{booktabs}
\usepackage{times,amsmath,amssymb,booktabs,tabulary,multirow,overpic,xcolor}
\definecolor{mygray}{gray}{0.6}
\definecolor{mygray-bg}{gray}{0.95}

\usepackage[pagebackref,breaklinks,colorlinks]{hyperref}

\usepackage[capitalize]{cleveref}
\crefname{section}{Sec.}{Secs.}
\Crefname{section}{Section}{Sections}
\Crefname{table}{Table}{Tables}
\crefname{table}{Tab.}{Tabs.}

\begin{document}

\title{Masked Contrastive Pre-Training for Efficient Video-Text Retrieval}

\author{  
  Fangxun Shu\textsuperscript{\rm 1} \quad Biaolong Chen\textsuperscript{\rm 1} \quad Yue Liao\textsuperscript{\rm 2} \quad Shuwen Xiao\textsuperscript{\rm 1} \quad Wenyu Sun\textsuperscript{\rm 1}  \\ Xiaobo Li\textsuperscript{\rm 1} \quad Yousong Zhu\textsuperscript{\rm 3} \quad Jinqiao Wang\textsuperscript{\rm 3}  \quad Si Liu\textsuperscript{\rm 2} \\
   \textsuperscript{\rm 1}Alibaba Group \quad
  \textsuperscript{\rm 2}Beihang University \quad
    \\ \textsuperscript{\rm 3}NLPR, Institute of Automation,
Chinese Academy of Sciences
 \\ {\tt\small \{shufangxun.sfx,biaolong.cbl\}@alibaba-inc.com}
}
\maketitle

\begin{abstract}
   We present a simple yet effective end-to-end Video-language Pre-training~(VidLP) framework, \textbf{Ma}sked \textbf{C}ontrastive Video-language Pretraining~(MAC), for video-text retrieval tasks. Our MAC aims to reduce video representation's spatial and temporal redundancy in the VidLP model by a mask sampling mechanism to improve pre-training efficiency. Comparing conventional temporal sparse sampling, we propose to randomly mask a high ratio of spatial regions and only feed visible regions into the encoder as sparse spatial sampling. Similarly, we adopt the mask sampling technique for text inputs for consistency. Instead of blindly applying the mask-then-prediction paradigm from MAE, we propose a masked-then-alignment paradigm for efficient video-text alignment. The motivation is that video-text retrieval tasks rely on high-level alignment rather than low-level reconstruction, and multimodal alignment with masked modeling encourages the model to learn a robust and general multimodal representation from incomplete and unstable inputs. Coupling these designs enables efficient end-to-end pre-training: reduce FLOPs (60\% off), accelerate pre-training (by 3$\times$), and improve performance. Our MAC achieves state-of-the-art results on various video-text retrieval datasets including MSR-VTT, DiDeMo, and ActivityNet. Our approach is omnivorous to input modalities. With minimal modifications, we achieve competitive results on image-text retrieval tasks.

\end{abstract}

\begin{figure}[t]
  \centering
  \includegraphics[width=1.0\linewidth]{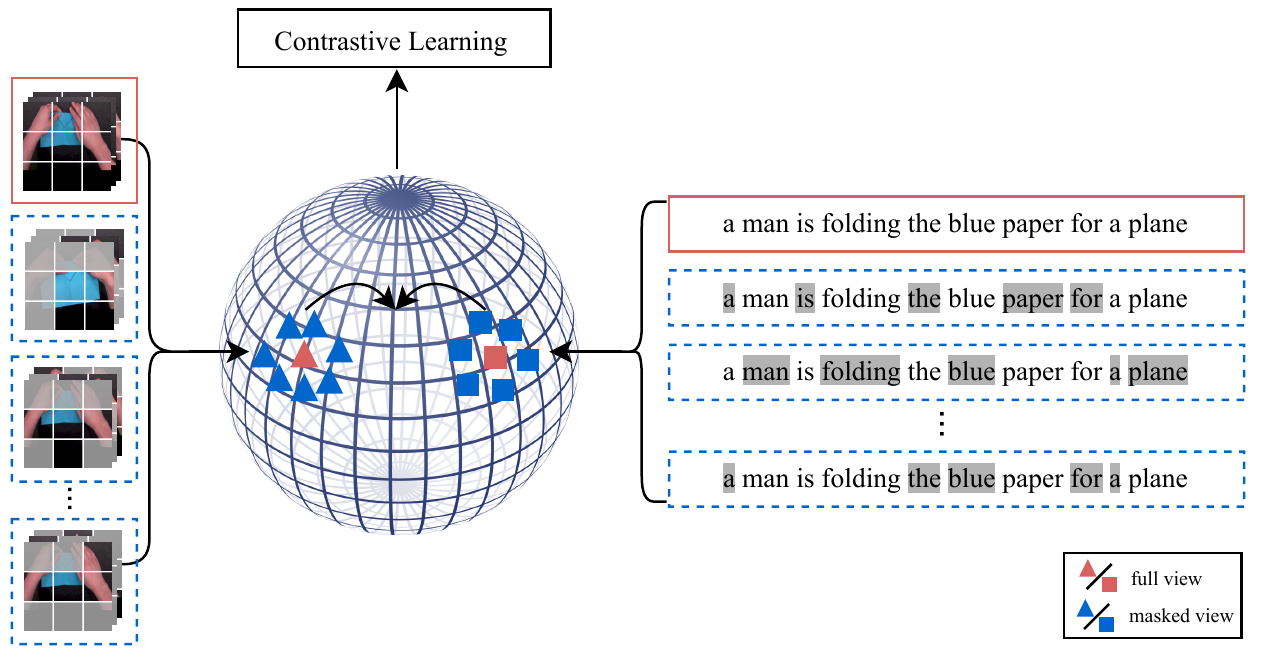}
   \caption{A brief illustration of our Masked Contrastive Pre-Training. The red samples are full view of input, and the blue samples are masked view of input. We generate random masked views of video and text and adopt a contrastive learning approach to learn the multimodal alignment.}
   \label{fig:brief intro}
\end{figure}

\vspace{-1mm}\section{Introduction}
\label{sec:intro}
Video-text retrieval is a fundamental multimodal understanding task aiming to compare a given text query with a set of video candidates to find the best-matching ones.
Thus, the core of video-text retrieval is how to align video and text into a unified space. 
With the development of computation resources, recent methods propose adopting Video-Language Pre-training techniques~(VidLP)~\cite{actbert, all-in-one, BridgeFormer, miles, UniVL, hero, Frozen, clipbert} to learn a feature alignment between the video and language modalities on large-scale datasets with a big model. 
Previous VidLP methods with a two-stage framework aim to leverage heavy pre-trained models (\emph{e.g.}, S3D~\cite{s3d}, SlowFast~\cite{slowfast} and VideoSwin~\cite{videoswin}) to extract fine-grained features to empower the encoder but suffer from huge computation costs for video processing, thus forced to break the VidLP into two isolated stages. 
However, the information translation is blocked by two isolated stages. 
Recent methods explore an end-to-end framework for VidLP, which designs an encoder with a series of cross-modal fusion modules to directly process video frames and text descriptions to obtain a multimodal representation. Despite the promising performance, such end-to-end VidLP methods require an encoder to process multiple full-resolution frames simultaneously for video spatio-temporal information extraction, which is computation-consuming. 
In this paper, we aim to explore an effective yet efficient VidLP mechanism for video-text retrieval.

We first investigate the efficiency improvement in VidLP. 
The key to efficient video-text pre-training lies in efficient video representation learning since videos have high temporal and spatial redundancy. Traditional end-to-end works mainly concentrate on sparse sampling techniques for video frames to reduce computations. However, such methods still directly process full-resolution video frames, so the spatial redundancy remains unsolved. Therefore, spatial redundancy reduction is an essential step in efficiency improvement. Recent mask sampling techniques in the visual domain~\cite{mae,videomae,st_mae} propose to randomly mask a high-ratio of spatial regions and adopt the unmasked regions to pre-train the encoder, which brings a new idea to reduce spatial redundancy. 
Inspired by this, we aim to introduce mask sampling methods into the VidLP framework to improve efficiency. Another core problem is how to learn effective video-text feature alignment with mask sampling. 
Therefore, the encoder is required to align video and text features into a unified space but keep their uni-modal semantic invariance based on unstable and incomplete masked inputs. 
We assume that cross-modal alignment with masking modeling clusters different masked views of the same sample into a stable space to learn a semantic invariance in the early stage, thus enabling robust and efficient cross-modal alignment.
Furthermore, visual mask sampling methods usually design a decoder to reconstruct mask regions, where the goal is to learn various low-level information. However, this is inconsistent with video-text alignment for high-level feature comparison. Thus blindly applying the masked-then-prediction diagram is negative. Instead, we adopt a masked contrastive learning approach as a pre-task to adapt to the downstream task, video-text retrieval.

To this end, we present a simple yet effective VidLP method, namely \textbf{Ma}sked \textbf{C}ontrastive Pre-Training (MAC), for video-text retrieval tasks. We perform video masking for spatio-temporal sparse sampling and text masking to make video and text complement each other and facilitate better video-text alignment. Different from ~\cite{mae, beit, videomae, st_mae}, we do not apply masked prediction on video and text but directly apply the cross-modal contrastive loss to pull video and text together. The final masking ratio is 60\% for video and 15\% for text, which is consistent with ~\cite{bert} and ~\cite{mae,videomae}. Our simple Masked Contrastive Pre-Training not only reduces the computation resource but also achieves state-of-the-art performance on various video-text retrieval tasks. Compared with the existing video-language pre-training methods, our MAC is efficient in multimodal alignment from two aspects. First, we significantly reduce the calculation and improve the pre-training speed by the proposed masking mechanism. Second, we greatly promote video-text alignment by the masked contrastive learning approach.

We report strong results on the video-text retrieval tasks. MAC achieves state-of-the-art results with less data, simpler pre-text tasks, and lighter architecture. For example, we achieve 38.9\% R@1 on MSRVTT while reducing FLOPs by 60\% and accelerating the 3 $\times$ pre-training. Our approach is also flexible and plug-and-play. For example, we achieve a competitive result on the image-text retrieval tasks without any strong data augmentations and multi-scale fusion designs. 

Our contributions can be summarized as follows: 1) We explore the masking mechanism in video-language pre-training. Instead of blindly applying the mask-then-prediction paradigm from MAE, we propose a masked-then-alignment paradigm, namely Masked Contrastive Pre-training, for efficient video-text alignment. 
2) We conduct experiments to show that the proposed masked contrastive learning is a better pre-text task than the masked prediction for video-text alignment. We also show that multimodal alignment with masked modeling encourages the model to learn not only cross-modal alignment but also uni-modal semantic invariance.
3) We outperform existing works on several video-text retrieval tasks with fewer FLOPs and faster training. We also achieve competitive results on image-text retrieval tasks, showing that MAC is flexible for various tasks.

\begin{figure*}[t]
\centering
\includegraphics[width=0.8\textwidth]{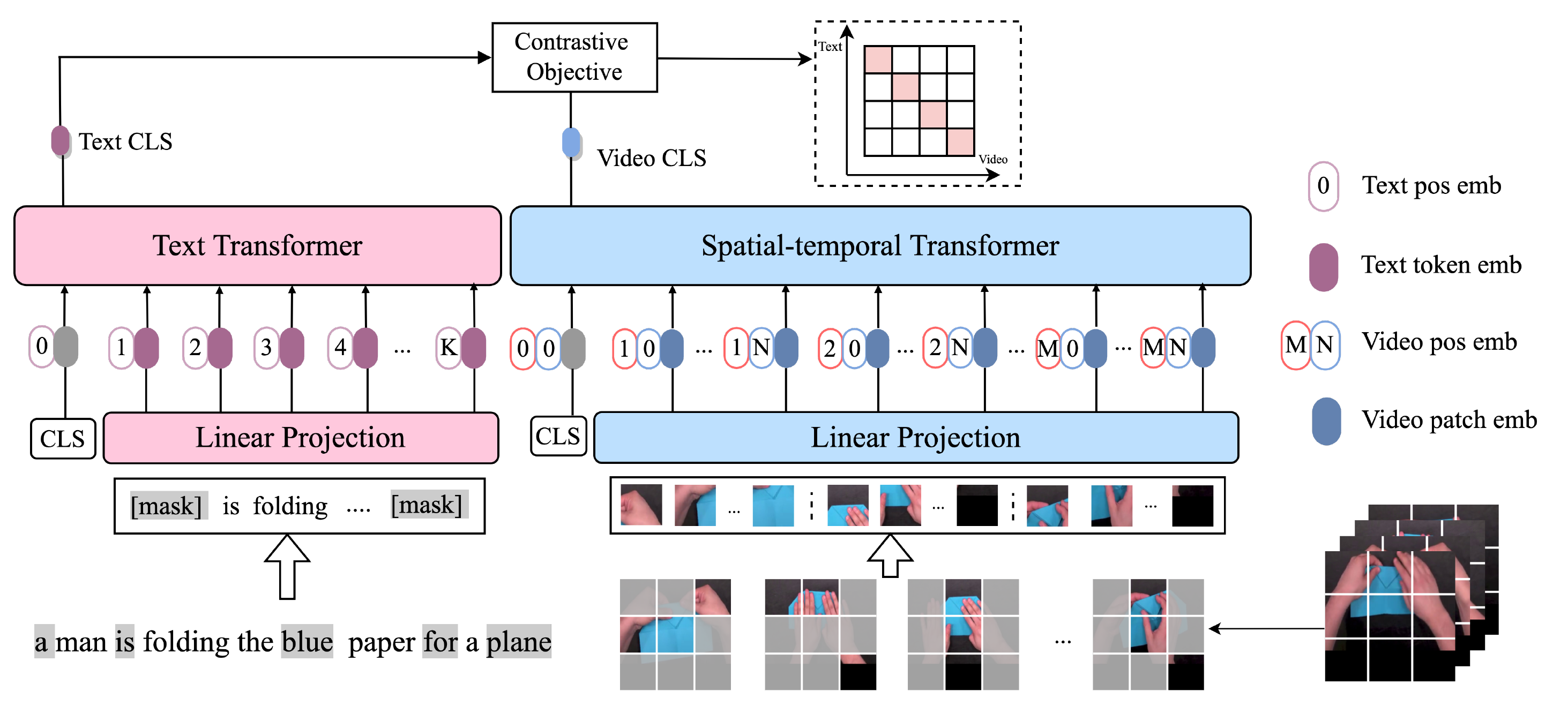}
\vspace{-2mm}
\caption{The framework of Masked Contrastive Pre-trainingfigures consists of a dual masking, a dual-stream encoder, and a contrastive objective. For video modality, we sample frames temporally and patches spatially. For text modality, we mask the whole words of the sentence. Then we feed the visible patches and words into the dual-stream encoder to pull the video and text together with a contrastive objective. }
\vspace{-2mm}
\label{fig:current_method}
\end{figure*}

\vspace{-1mm}\section{Related Works}
\label{sec:related works}

\noindent\textbf{Video-Text Alignment}
Video-text alignment aims to pull the paired video and text together in a unified space. Early works ~\cite{hero, UniVL} use uni-modal pre-trained models, such as video action recognition~\cite{slowfast, tsn, tsm, s3d} and image classification~\cite{vgg, resnet, swin} to extract multi-granularity features for performance improvement, which leads to a domain/task disconnection~\cite{clipbert}. Recently, end-to-end video-language pre-training combining large-scale datasets and pretext tasks has shown great potential. They usually design cross-fusion modules such as masked language modeling (MLM)~\cite{videobert, vl-bert}, video text matching (VTM) ~\cite{clipbert, actbert}, frame order modeling ~\cite{hero, merlot} and masked vision modeling ~\cite{hero, VIOLET}. Despite promising results, these designs increase computation and hinder end-to-end training. Considering the strong similarity between consecutive frames, ClipBERT ~\cite{clipbert}, and Frozen ~\cite{Frozen} propose sparse frame sampling to reduce temporal redundancy, enabling end-to-end training with raw video as input. This is becoming a popular trend with several follow-up works ~\cite{BridgeFormer, miles, alpro, OA-Trans}. However, such end-to-end VidLP methods still process full-resolution frames for video spatio-temporal information extraction, which is very computation-consuming. Inspired by ~\cite{mae, videomae, st_mae}, we perform mask sampling on video and text to achieve end-to-end video-text alignment.

\vspace{1mm}\noindent\textbf{Masked Multimodal Modeling}
Masked Modeling is one of the standard pretext tasks for pre-training. In the language domain, masked language modeling (MLM)~\cite{bert} predicts masked tokens of the input text, showing great generality on various downstream tasks. However, it cannot be easily extended to the vision domain due to the different properties between vision and language. Aligned with the success of ViT~\cite{vit, vivit}, visual input can be processed like language, making masked visual modeling (MVM) possible. BEiT~\cite{beit} and follow-up works~\cite{bevt, ibot, vimpac} utilize dVAE ~\cite{dvae} to encode visual patches into discrete semantic tokens, which can be trained in a BERT-style manner. MAE~\cite{mae} and follow-up works~\cite{simmim, videomae, st_mae} utilize the autoencoder to directly reconstruct RGB pixels in an end-to-end manner. It is natural to combine MLM and MVM in the multimodal domain~\cite{vl-beit, beit-3, multimae}. However, in the video-language domain, there exists limited work on masked modeling on both video and language. Violet~\cite{VIOLET} directly transfers BEiT to the video-language domain, which is a two-stage framework. Therefore, we propose end-to-end dual masking with contrastive learning, enabling efficient end-to-end video-language pre-training.

\vspace{-1mm}\section{Masked Contrastive Pre-Training}
\label{sec:method}
We propose \textbf{Ma}sked \textbf{C}ontrastive Pre-Training(\textbf{MAC}), a general video-language pre-training framework that enables efficient end-to-end multimodal alignment on the video-text retrieval tasks by performing dual masking on both video patches and text tokens. Different from ~\cite{mae, beit, videomae, st_mae}, we do not apply the masked prediction on video and text but directly apply the contrastive objective to pull the paired video and text together while pushing the unpaired video and text apart. \cref{fig:current_method} gives an overview of our MAC framework, which consists of the dual masking, the dual-stream encoder, and the cross-modal contrastive objective. We adopt a spatio-temporal sampling strategy on video, generating sparse patches with only a single or a few frames at each training step. We follow the BERT-wwm ~\cite{bert-www} to replace the whole words with \verb+[MASK]+. The visible sparse patches and text tokens are independently processed with the dual-stream encoder, followed by a contrastive loss to align the masked video and text in a shared feature space.

Formally, given the video-text pairs $(V, T)$, we first apply a video masking $\mathcal{M}^v$ and text masking $\mathcal{M}^t$ to get the masked video-text pairs $(\tilde{V}, \tilde{T})$, and then encoded with a video encoder $f_\theta$ and a text encoder $g_\theta$ to get the embedding of video $\{v_{cls},v_1,v_2,...,v_M \}$ and text $\{t_{cls},t_1,t_2,...,t_N \}$. A contrastive objective (\emph{i.e.}, infoNCE~\cite{cpc_infonce}) on the video \verb+[CLS]+ embedding $v_{cls}$ and text \verb+[CLS]+ embedding $t_{cls}$ is introduced to pull the paired video-text together and push the unpaired video-text apart. We will discuss the dual masking strategy in \cref{sec:dual masking}, the dual-stream encoder in \cref{sec:dual-stream encoder} and the masked contrastive objective in \cref{sec:contrastive objective}.

\vspace{-1mm}\subsection{Dual Masking Mechanism}\vspace{-1mm}
\label{sec:dual masking}
The spatio-temporal redundancy of video is the key to the efficiency of video-language pre-training. Existing works focus on temporal redundancy and reduce computation through temporal sparse sampling. However, they still feed full-resolution frames as input and suffer from spatial redundancy. With the ConvNet ~\cite{vgg, resnet} architecture, the spatial resolution of videos is usually downsampled (\emph{e.g.}, from 224 to 112), which results in a performance drop. Benefiting from the success of the transformer architecture in the computer vision domain, recent methods~\cite{mae, beit} draw inspiration from MLM and propose a masked visual modeling (MVM) technique to randomly mask a high portion of spatial patches and encode with visible patches, which greatly reduces spatial redundancy. Furthermore, encouraging the encoder to produce a better representation for downstream tasks. Inspired by this, we explore a joint video-language masking mechanism in the VidLP framework to perform video masking for spatio-temporal sparse sampling and text masking to make video and text complement each other and promote better video-text alignment.

\vspace{1mm}\noindent\textbf{Video Masking.} Formally, given a raw video with a resolution of $H$ $\times$ $W$ and consecutive frames, we first sparsely sample $M$ frames along the temporal dimension to reduce the temporal redundancy. Following the ~\cite{vit}, we divide each frame into $N$ non-overlapping patches in the spatial dimension to generate spatio-temporal patches $V \in \mathbb{R}^{M \times N \times P \times P \times C}$, and flatten by a linear projection layer to produce the spatial-temporal tokens, which can be processed like text tokens. To fully learn the relations between patches, we add learnable spatio-temporal positional embeddings $E_s \in \mathbb{R}^{N \times D} $ and $E_t \in \mathbb{R}^{M \times D}$ to encode the spatio-temporal position, where $M$ is the maximum number of input video frames, and $N$ is the maximum number of non-overlapping patches. We assign the patches along the same temporal dimensions with the same $E_s$ and the patches along the same spatial dimensions with $E_t$. 

\begin{figure}[t]
  \centering
  \includegraphics[width=0.9\linewidth]{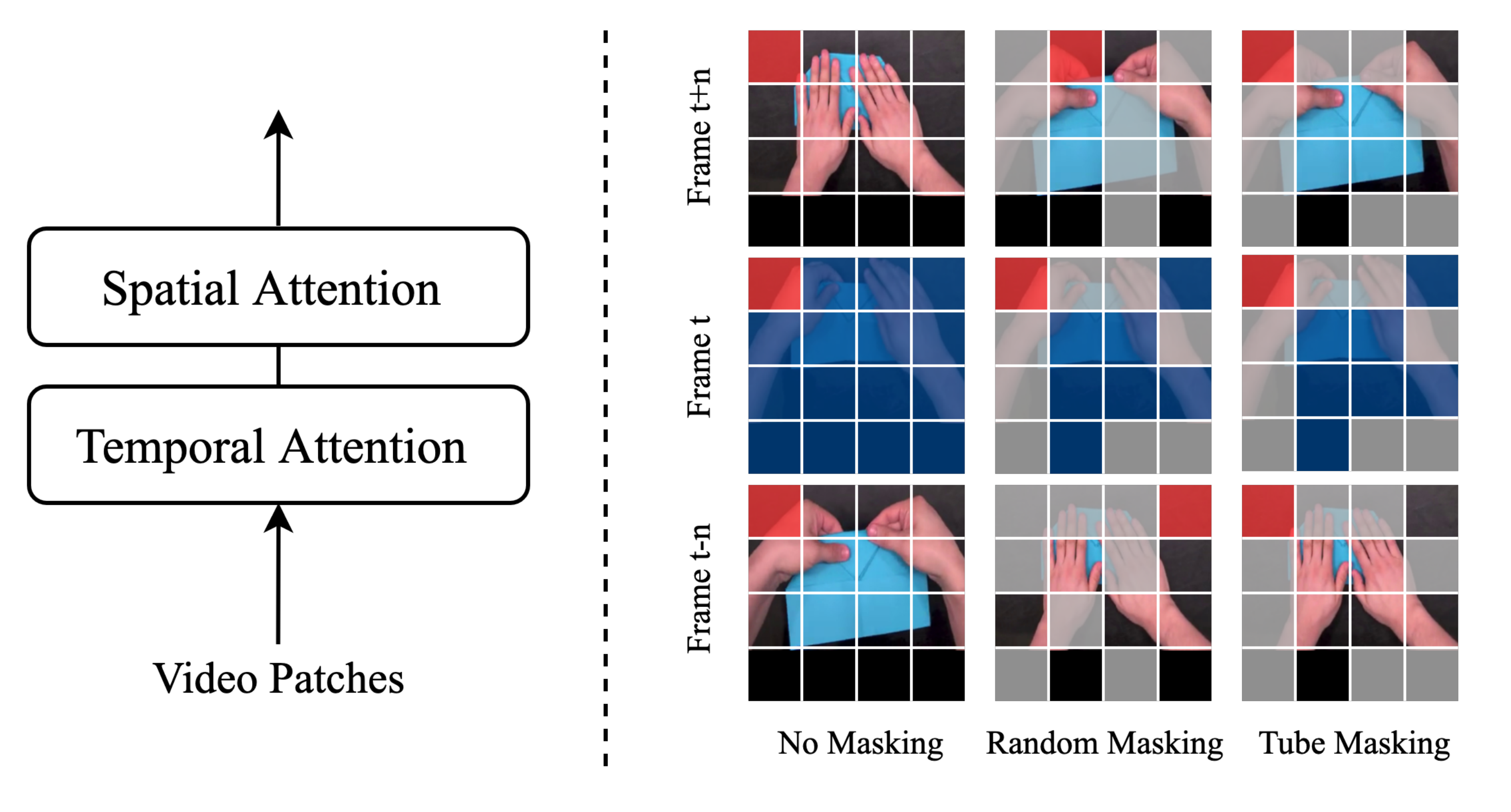}
   \caption{A illustration of divided space-time attention (left) and masking mechanisms (right), where \textbf{Grey} are masked patches, \textbf{Red} are spatial attention patches and \textbf{Blue} are temporal attention patches. Coupling random masking with divided space-time self-attention introduces more spatio-temporal diversity.}\vspace{-2mm}
   \label{fig:masking-type}
\end{figure}

We randomly mask a portion of patches with ratio $\rho_v$, and only the visible patches are available for the encoder. The high masking ratio significantly reduces the input tokens, thus reducing computation. From VideoMAE~\cite{videomae} and ST-MAE~\cite{st_mae} perspectives, the high masking ratio also prevents the model from simply copying neighborhood pixels for \textbf{low-level reconstruction} and use an extreme masking ratio of 90\%. However, our MAC aims to \textbf{high-level alignment}, and an extremely high ratio severely compromises alignment. Therefore, We achieve a balance between computation and performance by masking 60\% of video. As shown in \cref{fig:masking-type}, we experiment on different video masking strategy $\mathcal{M}_v$ such as random masking and tube masking, and find that only the proper masking with the proper network architecture can achieve its maximum potential. 

\vspace{1mm}\noindent\textbf{Text Masking.} For MLM in BERT~\cite{bert}, 80\% are replaced with \verb+[MASK]+, 10\% are randomly replaced, and 10\% remain unchanged. We perform minor modifications on BERT text masking mechanism by using only the \verb+[MASK]+ token to replace all the masked tokens to avoid random noise. Besides, we mask all tokens of the whole word to avoid information leakage of other tokens. 

\vspace{-1mm}\subsection{Dual-stream Encoder}\vspace{-1mm}
\label{sec:dual-stream encoder}
Video-language pre-training for video-text retrieval tasks can be classified as single-stream and dual-stream. The single-stream designs a cross-modal encoder for  cross interactions, which introduces the extra parameters and leads to heavy computation. Dual-stream only contains an independent video encoder and a text encoder, using contrastive loss to pull the video and text together, which is more efficient and flexible. Therefore, our MAC adopts the dual-stream architecture, consisting of a video encoder and a text encoder. This video encoder is transformer-based which can fully model spatial-temporal relationships. However, the computation complexity of the self-attention operation is $\mathcal{O}(n^2)$, we therefore adopt a divided space-time self-attention~\cite{vivit, timesformer} to reduce the dimension of attention matrix from $S \times T$ to $S+T$. The text encoder uses DistillBERT~\cite{distilbert} trained on the large-scale corpus. 

\vspace{1mm}\noindent\textbf{Video Encoder.} As mentioned above, we adopt a divided space-time self-attention architecture, which first performs attention in the temporal dimension, then in the spatial dimension. Combined with a proper masking mechanism, in addition to reducing the computation, it can also promote representation learning. As shown in the \cref{fig:masking-type}, no masking and tube masking only attend on patches at the same spatial location of different temporal locations. While random masking enables the attention of patches at different spatial locations of different temporal locations, thereby capturing a wider range of spatio-temporal diversity. 

\vspace{1mm}\noindent\textbf{Text Encoder.} We adopt BERT~\cite{bert} as the text encoder, which only operates on the visible text tokens $T_u$ to generate text embeddings. A learnable class token \verb+[CLS]+ is concatenated at the beginning of the text sequence, serving as the final representation of the whole text.

\vspace{-1mm}\subsection{Contrastive Objective}\vspace{-1mm}
\label{sec:contrastive objective}
Contrastive learning aims to learn a discriminative feature space where positive samples are close to each other while negative samples are far apart. In the video-language domain, the contrastive objective is formulated as follows:
\begin{equation}
\small
    d(f_\theta (v^+), g_\theta(t^+)) \ll d(f_\theta (v^{+/-}), f_\theta(t^{-/+}))
\end{equation}
It takes a pair of video-text and minimizes the embedding distance from the same pair, and maximizes the distance otherwise. Specifically, given a set of video-text pairs $(v,t)$, we directly apply infoNCE~\cite{cpc_infonce} on the pairs. The final training objective $\mathcal{L}_{V T C}$ is formulated as follows:

\begin{equation}
\small
\mathcal{L}_{V T C}=-\frac{1}{N}\sum_{i=1}^{N}\left[\mathcal{S}(v_i, t_i)+\mathcal{S}(t_i, v_i)\right]
\end{equation}

\begin{equation}
\small
\mathcal{S}\left(x_i, y_i\right)=-\log \frac{\exp \left(x_i^T y_i / \tau\right)} {\sum_{j=1}^N \exp \left(x_i^T y_j / \tau\right)}
\end{equation}
where $N$ is the batch size and $ \tau$ is the temperature hyperparameter.

\vspace{-1mm}\section{Experimental Setup}
In this section, we introduce the pre-training dataset and downstream tasks. We will describe our experimental setup in detail to ensure reproducibility.

\vspace{-1mm}\subsection{Pre-training Datasets}

We use WebVid-2M ~\cite{Frozen} with 2.5M video-text pairs. To improve the spatial relationship and accelerate temporal training. Google Conceptual Captioning (CC3M) ~\cite{cc3m} with 3.3M image-text pairs is also used. During the pre-training stage, we first randomly sample a single frame from Webvid2M and jointly train with CC3M for image-text alignment, then we train Webvid2M soley for video-text alignment separately. The number of pairs used in the pre-training is 5.5M for image-text pairs and 2.5M for video-text pairs, which is much smaller than the commonly used Howto100M ~\cite{howto100m} with 120M video-text pairs.  

\vspace{-1mm}\subsection{Downstream Tasks}

\noindent\textbf{Text-to-Video Retrieval.} 1) MSR-VTT ~\cite{msrvtt} contains 10k videos with a length that ranges from 10 to 32 seconds and 200k captions. We follow previous works ~\cite{Frozen, taco} to split into 9K and 1k videos for training and testing. 2) DiDeMo ~\cite{didemo} contains 10K videos with 40K sentences. Following [2], We concatenate all captions to perform paragraph-to-video retrieval. (3) ActivityNet ~\cite{activitynet} contains 20K videos with 100K captions. Following ~\cite{clipbert}, We concatenate all captions to perform paragraph-to-video retrieval. 4) MSVD ~\cite{msvd} contains 1.9k videos with a length that ranges from 1 to 62 seconds and 80K captions. The number of training, validation and testing videos is 1200, 100, and 670, respectively. We report the performance using standard retrieval metrics: recall at rank K (R@K, higher is better), median rank (MdR, lower is better).

\vspace{1mm}\noindent\textbf{Text-to-Image Retrieval.} Our video encoder is omnivorous to video and image, which can be seamlessly applied to image-text retrieval tasks and achieve competitive performance. We thereby evaluate on the image-text dataset Flickr30k ~\cite{flickr30k} , which contains 30k images with 5 captions per image. We follow the standard split of 29k, 1k and 1k for training, validation, and testing, respectively.

\begin{table*}[]
    \centering
    \small
        \begin{tabular}{ll|lll|cccc}
        \toprule
        \textbf{Method}       & \textbf{E2E} & \textbf{PT Dataset} & \textbf{\# Pairs}  & \textbf{\# Frames} & \textbf{R@1} $\uparrow$   & \textbf{R@5} $\uparrow$   & \textbf{R@10}  $\uparrow$  & \textbf{MedR}  $\downarrow$\\ \midrule
        ActBERT ~\cite{actbert}    &   & HowTo100M & 120M  & 32  & 8.6  & 23.4                     & 33.1      & 10.0              \\
        HERO~\cite{hero}  &   & HowTo100M          & 120M     & 32  & 16.8 & 43.4                     & 53.7       & -              \\
        CE~\cite{ce}        &   & -     & -    &  -  & 20.9  & 48.8 & 62.4     & 6.0              \\
        UniVL~\cite{UniVL}        &   & HowTo100M     & 120M    &  32  & 21.2 & 49.6                     & 63.1       & 6.0              \\
        ClipBERT~\cite{clipbert}    & $\checkmark$ & COCO,Visual Genome       & 5.6M   & 8    & 22.0 & 46.8             & 59.9     & 6.0               \\
        MMT ~\cite{mmt}       &   & HowTo100M          & 120M   &  32  & 26.6 & 57.1           & 69.6     & 4.0                \\
        TACo ~\cite{taco}        &   & HowTo100M   & 120M   &  48   & 28.4 & 57.8    & 71.2     & 4.0                \\
        SupportSet  ~\cite{support}        &   & HowTo100M          & 120M  &  -    & 30.1 & 58.5                     & 69.3     & 3.0                \\
        HiT  ~\cite{HiT}        &   & HowTo100M          & 120M    &  32 & 30.7 & 60.9  & 73.2     & 2.6                \\
        ALPRO  ~\cite{alpro}      &   & CC3M,WebVid2M       & 5.5M   &  8  & 33.9 & 60.7                     & 73.2   & 3.0                   \\
        Frozen~\cite{Frozen}       & $\checkmark$ & CC3M,WebVid2M        & 5.5M   &  4   & 34.2 & 61.1       & 71.6           & 3.0 \\
        VIOLET~\cite{VIOLET}      &   & CC3M,WebVid2M,YT180M      & 185.5M   &  4  & 34.5 & 63.0                     & 73.2  & -                   \\
        BridgeFormer~\cite{BridgeFormer} &   & CC3M,WebVid2M        & 5.5M   &   4   & 37.6 & 64.8                     & 75.1     & 3.0                \\
        MILES ~\cite{miles}       & $\checkmark$ & CC3M,WebVid2M       & 5.5M  &  4   & 37.7 & 63.6                     & 73.8    & 3.0                 \\
        All-in-one ~\cite{all-in-one}  & $\checkmark$ & HowTo100M,WebVid2M   & 122.5M  & 9 & 37.9 & \textbf{68.1}                     & \textbf{77.1}   & -                 \\
        \cellcolor{mygray-bg}Ours         & \cellcolor{mygray-bg}$\checkmark$ &\cellcolor{mygray-bg}CC3M,WebVid2M        &\cellcolor{mygray-bg}5.5M    &\cellcolor{mygray-bg}4   &\cellcolor{mygray-bg}\textbf{38.9} &\cellcolor{mygray-bg}63.1 &\cellcolor{mygray-bg}73.9 &\cellcolor{mygray-bg}3.0 \\ \bottomrule
        \end{tabular}\vspace{-2mm}
        
        \caption{Comparision with SOTA on text-to-video retrieval results for MSR-VTT, 1k-A. \textbf{E2E}: methods operate directly on raw video and text. \textbf{\# Pairs}: number of pairs for pre-training. \textbf{\# Frames}: number of frames for pre-training. \textbf{R@K}: recall at rank N, higher is better, \textbf{MedR}: median rank, lower is better}
        \label{tab:MSRVTT}
        \end{table*}

\begin{table*}[htb]
    \small
    \aboverulesep = 0.55mm
    \belowrulesep = 0.55mm
    \centering	
 \resizebox{1.0\textwidth}{!}
 {
    \begin{tabular}	{l | 
    l@{\hspace{1.5\tabcolsep}} l@{\hspace{1.5\tabcolsep}} l@{\hspace{1.5\tabcolsep}} |
    c@{\hspace{1.5\tabcolsep}} c@{\hspace{1.5\tabcolsep}} c@{\hspace{1.5\tabcolsep}}  
    c@{\hspace{1.5\tabcolsep}} c@{\hspace{1.5\tabcolsep}} c@{\hspace{1.5\tabcolsep}} 
    c@{\hspace{1.5\tabcolsep}} c@{\hspace{1.5\tabcolsep}} c@{\hspace{1.5\tabcolsep}} }
    
    \toprule {\textbf{Methods}}
    & \multicolumn{3}{c|}{\footnotesize\textbf{PT Config}}
    & \multicolumn{3}{c}{\footnotesize\textbf{ActivityNet(180s)}}  
    & \multicolumn{3}{c}{\footnotesize\textbf{DiDeMo(28s)}}
    & \multicolumn{3}{c}{\footnotesize\textbf{MSVD(10s)}} \\

        &  \footnotesize\textbf{Architecture}
        &  \footnotesize\textbf{\# Pairs}
        &  \footnotesize\textbf{\# Frames}
        &  \footnotesize\textbf{R@1}$\uparrow$ 
        & \footnotesize\textbf{ R@5}$\uparrow$
        &  \footnotesize\textbf{R@10}$\uparrow$ 
        & \footnotesize\textbf{R@1}$\uparrow$
        & \footnotesize \textbf{R@5}$\uparrow$ 
        & \footnotesize\textbf{R@10}$\uparrow$
        & \footnotesize\textbf{R@1}$\uparrow$
        & \footnotesize \textbf{R@5}$\uparrow$ 
        & \footnotesize\textbf{R@10}$\uparrow$ \\ \midrule
        
        CE~\cite{ce}  & V,T,V$^{expert}$ & - & - & 18.2 & 47.4 & - & 16.1 & 41.1 & -  & 19.8 & 49.0 & 63.8 \\
        ClipBERT~\cite{clipbert}  & V,T,C & 5.6M & 8 & 21.3 & 49.0 & 63.5 & 20.4 & 48.0 & 60.8  & - & - & - \\
        Frozen~\cite{Frozen} & V,T & 5.5M & 4 & 27.6 & 57.4 & 71.9 & 31.0 & 59.8 & 72.4 & 33.7 & 64.7 & 76.3  \\
        All-in-one~\cite{Frozen} & U & 122.5M & 9 & 22.4 & 53.7 & 67.7 & 32.7 & 61.4 & 73.5 & - & - & -  \\  
        TACo~\cite{taco}  & V,T,C & 120M & 48 & 25.8 & 56.3& 93.8 & - & - & - & - & - & -  \\
        HD-VILA  & V,T,C & 103M & 14 & 28.5 & 57.4 & 94.0 & 28.8 & 57.4 & 69.1 & - & - & -  \\
        OA-Trans~\cite{OA-Trans} &   V,T,V$^{det}$ & 5,5M &  8  & - & - & - &   34.8    &  64.4  & 75.1   &  39.1 & \textbf{68.4} & \textbf{80.3}       \\
        ALPRO~\cite{alpro}    &   V,T,C,V$^{pem}$  & 5.5M & 8  & - & - & - &   35.9         &  \textbf{67.5}       & \textbf{78.8}         & - & - & -\\
        BridgeFormer~\cite{BridgeFormer}  &   V,T,V$^{mcq}$  & 5.5M &  8 & - &  -  & - &   37.0   &  62.2  & 73.9  & - & - & -  \\
        MILES~\cite{BridgeFormer}    &   V,T,V$^{ema}$  &  5.5M & 8 & - &  -  & - &   36.6         &  63.9      & 74.0       & - &  -  & - \\
        MMT~\cite{mmt} & V,T,C & 120M & 32 & 28.7 & 61.4 & - & - & - & - & - & - & -  \\
        SupportSet~\cite{support} & V,T,C & 120M & - & 29.2 & 61.6 & - & - & - & - & 28.4 & 60.0 & 72.9  \\
        HiT~\cite{HiT} & V,T,V$^{ema}$ & 120M & 32 & 29.6 & 60.7 & - & - & - & - & - & - & -  \\
        \cellcolor{mygray-bg}\cellcolor{mygray-bg}Ours &\cellcolor{mygray-bg}V,T & \cellcolor{mygray-bg}5.5M &\cellcolor{mygray-bg}4 &\cellcolor{mygray-bg}\textbf{30.6} &\cellcolor{mygray-bg}60.5 &\cellcolor{mygray-bg}74.7 &\cellcolor{mygray-bg}36.7 &\cellcolor{mygray-bg}64.5 &\cellcolor{mygray-bg}75.3 &\cellcolor{mygray-bg}36.9 & \cellcolor{mygray-bg}65.1 &\cellcolor{mygray-bg}75.7 \\
        \cellcolor{mygray-bg}Ours & \cellcolor{mygray-bg}V,T & \cellcolor{mygray-bg}5.5M & \cellcolor{mygray-bg}8 & \cellcolor{mygray-bg}\textbf{34.4} & \cellcolor{mygray-bg}\textbf{64.8} & \cellcolor{mygray-bg}\textbf{78.3} & \cellcolor{mygray-bg}\textbf{38.1} & \cellcolor{mygray-bg}65.7 & \cellcolor{mygray-bg}76.8 & \cellcolor{mygray-bg}\textbf{39.3} & \cellcolor{mygray-bg}67.5 & \cellcolor{mygray-bg}79.4\\     
        
        \bottomrule
    \end{tabular}
	}\vspace{-2mm}
    \caption
    {Text-to-video retrieval results on ActivityNet(180s), DiDeMo(28s) and MSVD(10s) with various video duration. \textbf{Architecture}: $V$, $T$, and $C$ are video, text, and cross-modal encoder, respectively. $U$ is a uni-encoder for video and text. $V^{expert}$ are a set of experts for feature aggregation. $V^{det}$ is a detector for ROI features. $V^{pem}$ is an encoder for prompt engineering. $V^{mcq}$ is an encoder for the pretext task MCQ. $V^{ema}$ is an encoder for EMA updating. \textbf{\# Pairs} and \textbf{\# Frames} are the number of pairs and frames for pre-training. 
     }\vspace{-2mm}
    \label{table: i21k-activitynet-didemo-msvd}

\end{table*}

\vspace{-1mm}\subsection{Implementation Details}
\noindent\textbf{Architecture.} Our architecture is dual-stream encoder, containing a video and text encoder. For the video encoder, we adopt a divided space-time attention architecture ~\cite{timesformer, vivit}, which consists of 12 space-time attention blocks with patch size $P$=16. We initialize the spatial blocks with ViT-B/16 ~\cite{vit} pretrained on the ImageNet-21k ~\cite{imagenet}. For the text encoder, we use DistilBERT ~\cite{distilbert}, which is trained on Wikipedia and Toronto Book Corpus. We use a linear head to project the video embedding and text embedding into a feature space of 256 dimensions. 

\vspace{1mm}\noindent\textbf{Masking the input.} For video, we first randomly sample $F$ frames, where each frame is divided into $N$ patches of size 16$\times$16 and randomly mask a portion of patches in each frame. We also experiment several masking strategies such as tube masking, which masks patches at the same spatial location across all frames. For text, we replace the whole words with \verb+[MASK]+. 

\vspace{1mm}\noindent\textbf{Training details.} Our work is built upon Pytorch. The video is resized to 224$\times$224 and augmented with random crop and horizontal flip. The masking ratios for video and text are 60\% and 15\%. During pre-training, we randomly sample a single frame from Webvid2M and jointly train with CC3M to focus on spatial alignment, then randomly sample 4 frames from Webvid2M to focus on temporal alignment. The spatial alignment with 1 frame takes 20 epochs with a total batch size of 2048 and a learning rate of 1$\mathrm{e}$-4. The temporal alignment with 4 frames takes only 2 epochs with a batch size of 1024 and a learning rate of 3$\mathrm{e}$-5. We use the Adam optimizer ~\cite{adam} and decay the learning rate with a step schedule. The pre-training takes about 15 hours on 32 Tesla V100 GPUs. During finetuning, we uniformly sample 4 frames with a batch size of 128 and a learning rate of 3$\mathrm{e}$-5. For comparison with existing works, we sample 8 frames for DiDeMo, MSVD and ActivityNet.

\vspace{-1mm}\section{Results}
In \cref{sec:sota result}, we evaluate the effectiveness and generality of our method on various video-text retrieval tasks in terms of computation efficiency and performance. For a fair comparison, we reproduce Frozen ~\cite{Frozen}(34.2\% R@1) as a baseline with the same configuration, which is better than the official claimed 31.0\% R@1. We also found that the results of zero-shot and finetuning on MSRVTT are not consistent, so we only report the finetuning results. In \cref{sec:abalations}, we study the components of the proposed MAC, showing that masked contrastive pre-training is an effective framework for video-text alignment.  

\vspace{-1mm}\subsection{Comparison to State of the Art}
\label{sec:sota result}
\noindent\textbf{Text-to-Video Retrieval.} \cref{tab:MSRVTT} and \cref{table: i21k-activitynet-didemo-msvd} show the results on various video-text retrieval datasets. MAC achieves the SOTA performance in an efficient manner: \textbf{Smaller input}. We only use 4 frames for pre-training due to the efficient spatio-temporal modeling, while MMT~\cite{taco} uses 48 frames and OA-Trans~\cite{OA-Trans} use 8 frames. \textbf{Fewer computation}. With video masking, we significantly reduce 60\% FLOPs, thus takes 32 V100 GPUs with 15 hours for pre-training on 5M pairs, while OA-Trans takes 64 A100 GPUs with 5 days and Bridgeformer ~\cite{BridgeFormer} takes 40 A100 GPUs with 25 hours. \textbf{Lighter architecture}. We only use the dual-stream encoder architecture and contrastive learning to align video and text, while BridgeFormer~\cite{BridgeFormer} and MILES~\cite{miles} introduce modules with extra parameters for cross-modal fusion. We further evaluate the datasets with various video duration, including MSVD (10s), MSRVTT(15s), DiDeMo (28s), and ActivityNet(180s). Experimental results show that MAC can also fully capture the spatio-temporal relationships between video and text and outperform existing works by a large margin. Especially, the good results on ActivityNet indicate that our approach can generalize well to long videos.

\begin{table}[]
    \centering
    \resizebox{0.48\textwidth}{!}{
    \small
    \begin{tabular}{l|l|lll}
    \toprule
    \textbf{Method}   & \textbf{PT Dataset}& \textbf{R@1} $\uparrow$  & \textbf{R@5} $\uparrow$   & \textbf{R@10}  $\uparrow$ \\ \midrule
    SGRAF~\cite{scaraf}   &  VG  & 58.5 & 83.0 & 88.8  \\
    ViLT~\cite{vilt}    & CC3M,VG,COCO,SBU & 64.4 & 88.7 & 93.8   \\
    UNITER~\cite{uniter}  & CC3M,VG,COCO,SBU & 75.6 & 94.1 & 96.8  \\
    \cellcolor{mygray-bg}Ours    & \cellcolor{mygray-bg}CC3M,WV2M   & \cellcolor{mygray-bg}\textbf{79.3} &  \cellcolor{mygray-bg}\textbf{94.7}    &    \cellcolor{mygray-bg}\textbf{97.2}      \\ \bottomrule
    \end{tabular}
    }
    \caption{Text-to-image retrieval results on Flickr30k. Our approach achieves a competitive result with custom designs such strong as data augmentations and multi-scale fusion.}
    \label{tab:Flickr30k}
    \end{table}

\begin{table}[]
\centering
\resizebox{0.48\textwidth}{!}{
\small
\begin{tabular}{l|ll|ccc}
\toprule
\textbf{Method} & \textbf{Params(M)} & \textbf{FLOPs(G)} & \textbf{R@1} $\uparrow$  & \textbf{R@5} $\uparrow$   & \textbf{R@10}  $\uparrow$\\ \midrule
Frozen~\cite{Frozen}   & 180.7    & 189.3  & 34.2 & 61.1  & 71.6 \\
VIOLET~\cite{VIOLET}   &  196.7    & 191.1 & 34.5 & 63.0  & 73.2 \\
BridgeFormer~\cite{BridgeFormer}    & 266.0  & 286.1  & 37.6 & \textbf{64.8}  & \textbf{75.1} \\
MILES~\cite{miles}    & 295.5  & 367.5  & 37.7 & 63.6   & 73.8 \\
\cellcolor{mygray-bg}\textbf{Ours}    &\cellcolor{mygray-bg}\textbf{180.7} &\cellcolor{mygray-bg}\textbf{83.3}    &\cellcolor{mygray-bg}\textbf{38.9} &\cellcolor{mygray-bg}63.1 &\cellcolor{mygray-bg}73.9 \\ 
\bottomrule
\end{tabular}
}
\caption{Comparison with Params and FLOPs during pre-training. The listed works are all based on the dual-stream architecture. We use 4 frames with 224 resolution for video and 128 of text length. }
\label{tab:flops and params}
\end{table}

\begin{figure}[t]
  \centering
  \includegraphics[width=0.8\linewidth]{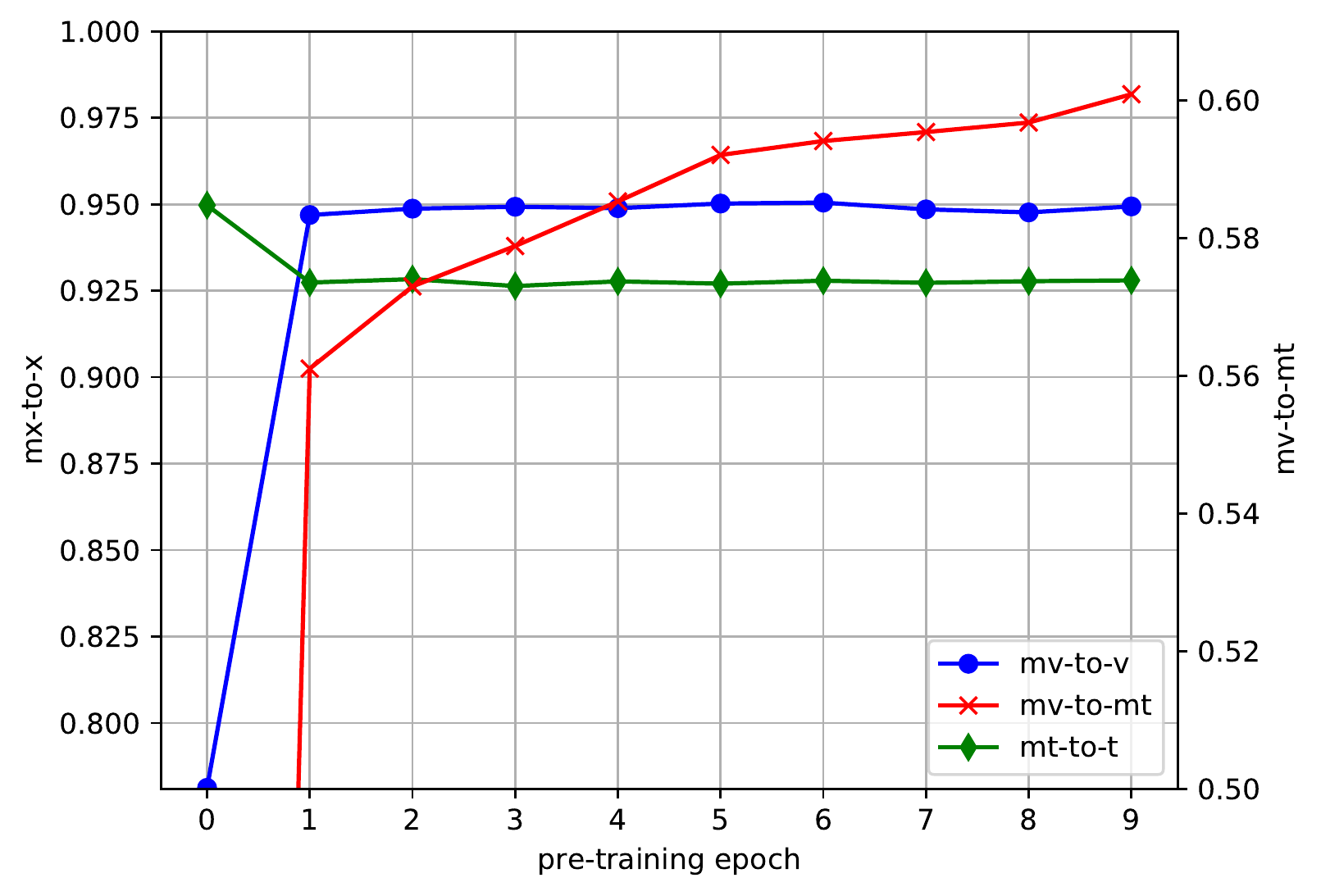}
   \caption{Similarity during pre-training. \textbf{Blue}: Video similarity between masked views and full view. \textbf{Green}: Text similarity between masked views and full view. \textbf{Red}: Similarity between masked video and masked text.}
   \label{fig:similarity}
\end{figure}

\vspace{1mm}\noindent\textbf{Text-to-Image Retrieval.} Our approach adapts to both image and video by treating the image as a single frame video. Therefore, it is flexible for image-text retrieval tasks. We thus conduct experiments on the Flickr30K~\cite{flickr30k} dataset. \cref{tab:Flickr30k} shows that our masked contrastive learning also achieves competitive results without custom designs such as strong augmentation (\emph{e.g.}, RandAugment ~\cite{randaugment}) and multi-scale feature fusion. This suggests that our approach is effective in both image and video domains and can be flexible plug-and-played into the existing multimodal frameworks.

\vspace{1mm}\noindent\textbf{Efficient Masked Alignment.} Most contrastive methods rely on heavily strong data augmentation techniques \cite{randaugment, cutout} to significantly change the visual appearance but keep the semantic meaning unchanged. However, in our experiment, we achieve promising performance with basic random flip and crop. We find the key is masking. Multimodal alignment with masking modeling is a difficult pretext task. The encoder must align video and text features into a unified space but keep their uni-modal semantic invariance based on unstable and incomplete masked inputs. 

To verify our assumption, we conduct experiments to analyze masked cross-modal alignment trend. We randomly mask a set of origin videos 100 times and compute the similarity between all masked views and the corresponding full view. As shown in \cref{fig:similarity}. The pre-training focus on uni-modal representation learning in the early stage (about 1 epoch), then on cross-modal alignment. This is consistent with our assumption in \cref{sec:intro} that multimodal alignment with masking modeling clusters different masked views of the same sample into a stable space in the early stage to learn semantic invariance, thus enabling more efficient and robust cross-modal alignment.

\vspace{1mm}\noindent\textbf{Params and FLOPs.} We evaluate Params and FLOPs on several works built on the dual-stream architecture. We feed a video with 4 frames wth 224$\times$224 resolution and text with length of 128 as default input. The results in \cref{tab:flops and params} show that our approach achieves the best result. VIOLET and BridgeFormer all introduce modules with extra parameters to enhance the interaction between video and text. However, we add no modules and only perform masking to reduce the FLOPs from 189.3G to 83.3G with a video masking ratio of 60\%.


\vspace{-1mm}\subsection{Ablation Studies}\vspace{-1mm}
\label{sec:abalations}
In this section, we systematically study the designed components of MAC. We shortly pre-train 6 epochs (5 for image-text and 1 for video-text) and finetune 50 epochs on MSRVTT. Unless otherwise specified, the default video masking ratio and text masking ratio are 60\% and 15\%.

\vspace{1mm}\noindent\textbf{Pretext Task Design.} The mainstream masked modeling works predict the masked part (e.g., MLM and MIM), which is a low-level reconstruction pretext task. We aim to pre-training for video-text retrieval tasks, which require high-level alignment. This suggests that the masked-then-prediction paradigm in MAE~\cite{mae} is inconsistent with the goal of retrieval tasks. We, therefore, propose the masked-then-alignment paradigm, aligning the masked views directly. As shown in \cref{tab:task-ablations}, we conduct detailed experiments on various pre-text tasks to verify our assumption: 1) VTC, 2) VTC+MLM, 3) VTC+LM, 4) VTC+MVP+LM. For a fair comparison, the coefficient of all tasks to the total loss is set to 1. It can be seen that adding generative pretext tasks results in a performance drop, which is consistent with our assumption that multimodal alignment requires high-level semantic understanding. Besides, it adds an extra decoder and is difficult to optimize, resulting in high training overhead. We also find that the ranking of performance drop is LM $>$ MLM $>$ MVM, showing that language reconstruction is more difficult than visual reconstruction. A possible explanation is that vision is continuous while language is discrete, which makes language reconstruction more difficult, thus forcing the model to focus on uni-modal reconstruction rather than multimodal alignment. Recent works also explore the balance between high-level  understanding and low-level reconstruction, such as selecting  semantic targets~\cite{beit, maskft} rather than pixels. In the video-language domain, VIOLET ~\cite{VIOLET} reconstructs  visual tokens generated by dVAE ~\cite{dvae}. Although the promising performance of 34.5\% R@1, it still is lower than our Masked Alignment of 38.9\% R@1.

However, what level of semantics is required is an open question. It mainly depends on the downstream tasks, we assume that masked alignment is suitable for high-level semantic understanding (\emph{e.g.}, cross-modal retrieval), while masked prediction is suitable for subtitle generation or low-level semantic understanding (\emph{e.g.}, segmentation).

\begin{table}[h]
\centering
\small
\begin{tabular}{@{}cccc|ccc@{}}
\toprule
\textbf{VTC}  &  \textbf{MVM} & \textbf{MLM} & \textbf{LM} &  \textbf{R@1} $\uparrow$  & \textbf{R@5} $\uparrow$   & \textbf{R@10}  $\uparrow$   \\ \midrule
$\checkmark$ &                  &      &             & \textbf{36.4}      & \textbf{63.8}      & \textbf{73.0}    \\
$\checkmark$ & $\checkmark$     &      &          & 34.4  & 60.0  & 71.1         \\
$\checkmark$  &    &   $\checkmark$  &           &32.1 &  58.5 &  69.4         \\
$\checkmark$  &    &     &      $\checkmark$     &27.2 &  51.3 &  63.9         \\
$\checkmark$ & $\checkmark$ &    &  $\checkmark$   &26.3 &  51.3 & 64.0      \\ \bottomrule
\end{tabular}
\caption{Ablation experiments on task design. \textbf{VTC} directly aligns video and text via a video-text contrastive loss. \textbf{MVM} adopts a video decoder for Masked Video Prediction. \textbf{MLM} and \textbf{LM} are Masked Language Modeling and Language Modeling respectively }
\label{tab:task-ablations}
\end{table}

\vspace{1mm}\noindent\textbf{Video Masking Strategy.} As mentioned in \cref{sec:dual-stream encoder}, we adopt the divided space-time self-attention where the temporal attention and spatial attention are applied in turn. Combined with a proper masking mechanism, in addition to reducing the computation, we can also promote attention. As shown in the \cref{fig:masking-type}, no masking and tube masking only pay attention to patches at the same spatial location of different temporal locations. However, random masking enables the attention of patches at different spatial locations of different temporal locations, thereby capturing a wider range of spatio-temporal diversity. We conduct experiments to compare different video masking strategies, showing that cooperating random masking and divided space-time self-attention can achieve better results.

\begin{table}[h]
    \centering
    \small
    \begin{tabular}{@{}l|lll@{}}
    \toprule
    \textbf{Masking Strategy}  & \textbf{R@1} $\uparrow$  & \textbf{R@5} $\uparrow$   & \textbf{R@10}  $\uparrow$   \\ \midrule
    w/o    & 34.0 & 61.7 & 71.9 \\
    Tube  &   35.7     &  62.1      & 71.2       \\ 
    Random    &  \textbf{36.4}  &    \textbf{63.8}    &    \textbf{73.0}    \\ \bottomrule
    \end{tabular}
    \caption{Ablation experiments on different masking strategy on divided space-time attention. w/o: full video without any masking. Tube: mask same spatial locations across different temporal locations. Random: mask different spatial locations of different temporal locations. Combined with divided space-time attention, random masking performs better than tube masking.}
    \vspace{-2mm}
    \label{tab:maskingtype-ablations}
    \end{table}

\begin{figure}[htbp]
\centering
\setlength{\abovecaptionskip}{0.cm}
\begin{minipage}[t]{0.48\textwidth}
\centering
\includegraphics[width=8cm]{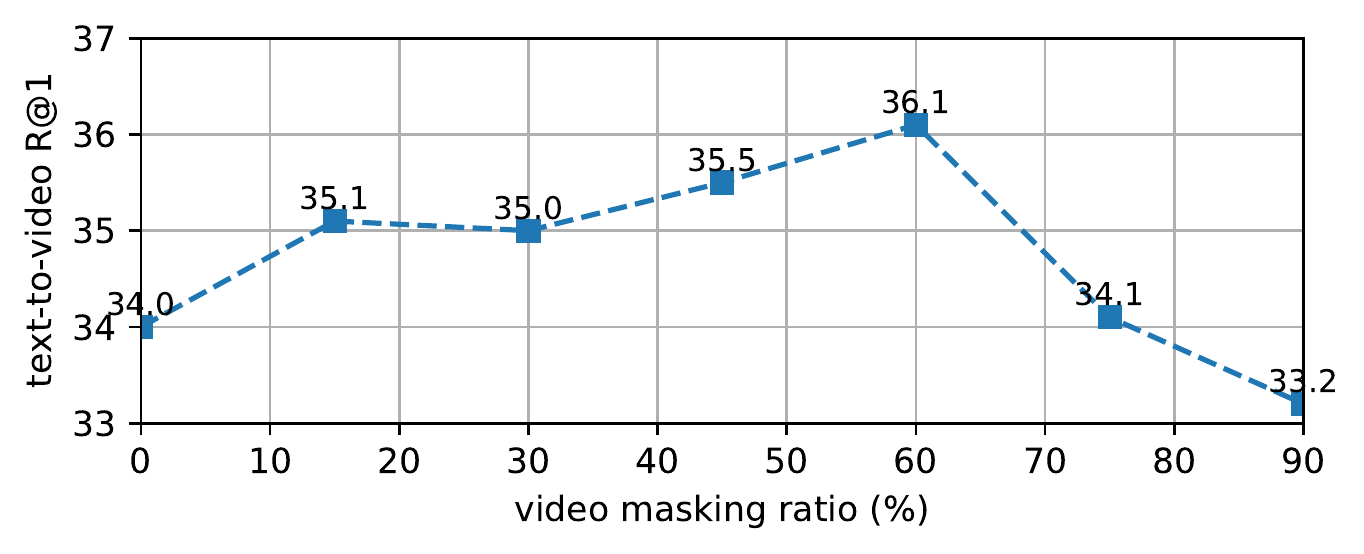}
\end{minipage}
\begin{minipage}[t]{0.48\textwidth}
\centering
\includegraphics[width=8cm]{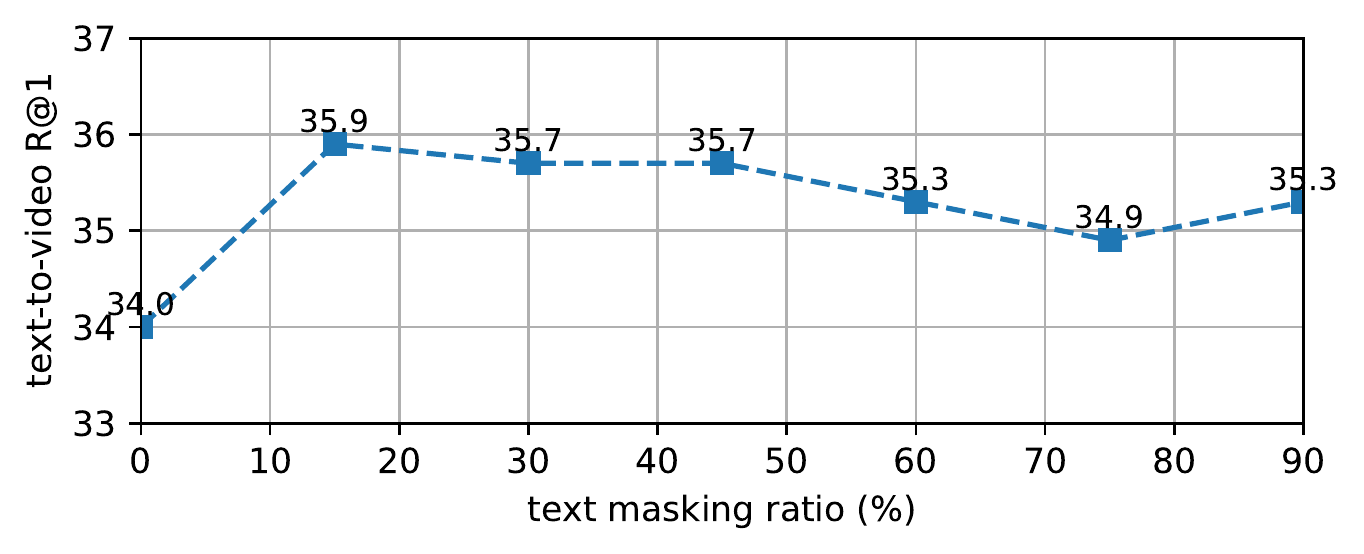}
 \caption{Ablation studies on masking ratio. We hold one modality unmasked when the other modality is masked. The masking ratio of video is relatively higher than that of text} 
\label{fig:masking ratio}\vspace{-1mm}

\end{minipage}
\end{figure}

\vspace{1mm}\noindent\textbf{Masking Ratio and Modality.} \cref{fig:masking ratio} shows the effect of masking ratio To quantitatively analyze the masking ratio, we hold one modality unmasked when the other modality is masked. It shows a high video masking ratio and low text masking ratio is a better combination for downstream tasks. This is consistent with the conclusions of MAE\cite{mae} and BERT~\cite{beit}: the information density of image is low, and the information density of text is high. We also found that under the multimodal setting, the video masking ratio is slightly lower than the visual domain, we assume that the high ratio leads to the large missing of visual entities, which makes multimodal alignment fail. Overall, an appropriate masking ratio can achieve a win-win situation in terms of computation and performance.

\cref{tab:modality-ablations} shows the effects of Dual Masking (masked video-text) and Single Masking (masked video/text). It can be seen that compared to the baseline without masking, both single masking and dual masking can enhance cross-modal alignment and improve the performance of downstream tasks. Besides, compared with single masking, dual masking performs mask sampling on both video and text, which increases the difficulty of cross-modal alignment and allows the model to learn more useful representations. Thus, we choose to mask both modalities simultaneously.

\begin{table}[h]
    \centering
    \small
    \begin{tabular}{@{}ll|lll@{}}
    \toprule
    \textbf{Video}  &  \textbf{Text}   &  \textbf{R@1} $\uparrow$  & \textbf{R@5} $\uparrow$   & \textbf{R@10}  $\uparrow$   \\ \midrule
                  &                  &  34.0      & 61.7      & 71.9    \\
    $\checkmark$  &                  & 35.8       & 62.2      & 72.2    \\ 
                  & $\checkmark$     & 35.6       & 60.0      & 72.0         \\  
    $\checkmark$  &  $\checkmark$    & \textbf{36.4}       &  \textbf{63.8}     & \textbf{73.0}         \\ \bottomrule 
    \end{tabular}
    \caption{Ablation experiments on masked modality. $\checkmark$ means the corresponding modality input is masked}
    \label{tab:modality-ablations}\vspace{-1mm}
    \end{table}

\vspace{-1mm}\section{Conclusion}\vspace{-1mm}
In this paper, we propose an end-to-end video-text pre-training framework for efficient video-text alignment, namely Masked Contrastive Pre-Training. Without blindly applying mask-then-prediction paradigm from MAE, we explore the masking contrastive mechanism based on the video language domain, and propose a mask-then-alignment paradigm  to efficiently learn a multimodal alignment. The experimental results show that our method can reduce 60\% of FLOPs, while improving the pre-training efficiency by 3 $\times$, and finally achieve SOTA performance on various video-text retrieval datasets.

{\small
\bibliographystyle{ieee_fullname}
\bibliography{egbib}
}
\end{document}